\documentclass{article}

\usepackage{microtype}
\usepackage{graphicx}
\usepackage{subcaption}
\usepackage{booktabs} 

\usepackage{hyperref}

\usepackage[preprint]{icml2026}

\usepackage{amsmath}
\usepackage{amssymb}
\usepackage{mathtools}
\usepackage{amsthm}

\usepackage{color}

\usepackage[capitalize,noabbrev]{cleveref}

\theoremstyle{plain}

\theoremstyle{definition}

\theoremstyle{remark}

\usepackage[textsize=tiny]{todonotes}

\usepackage{multirow}
\usepackage{multicol}
\usepackage{listings}
\usepackage{xcolor}
\usepackage{commands}

\lstdefinestyle{sysprompt}{
  basicstyle=\ttfamily\small,
  columns=fullflexible,
  breaklines=true,
  breakatwhitespace=false,
  keepspaces=true,
  showstringspaces=false,
  frame=single,
  rulecolor=\color{black!20},
  captionpos=b,
  abovecaptionskip=0.5em,
  belowcaptionskip=0.5em,
}

\icmltitlerunning{Physics-guided LLM Agent for Equation Discovery}

\begin{document}

\twocolumn[
  \icmltitle{Think like a Scientist: Physics-guided LLM Agent for Equation Discovery}



  \icmlsetsymbol{equal}{*}

  \begin{icmlauthorlist}
    \icmlauthor{Jianke Yang}{ucsd}
    \icmlauthor{Ohm Venkatachalam}{ucsd}
    \icmlauthor{Mohammad Kianezhad}{ucsd}
    \icmlauthor{Sharvaree Vadgama}{ucsd}
    \icmlauthor{Rose Yu}{ucsd}
  \end{icmlauthorlist}

  \icmlaffiliation{ucsd}{UCSD}

  \icmlcorrespondingauthor{Jianke Yang}{jiy065@ucsd.edu}
  \icmlcorrespondingauthor{Rose Yu}{roseyu@ucsd.edu}


  \vskip 0.3in
]



\printAffiliationsAndNotice{}  

\begin{abstract}
  Explaining observed phenomena through symbolic, interpretable formulas is a fundamental goal of science. Recently, large language models (LLMs) have emerged as promising tools for symbolic equation discovery, owing to their broad domain knowledge and strong reasoning capabilities. However, most existing LLM-based systems try to guess equations directly from data, without modeling the multi-step reasoning process that scientists often follow: first inferring physical properties such as symmetries, then using these as priors to restrict the space of candidate equations.
We introduce
\textbf{KeplerAgent}, an agentic framework that explicitly follows this scientific reasoning process. The agent coordinates physics-based tools to extract intermediate structure and uses these results to configure symbolic regression engines such as PySINDy and PySR, including their function libraries and structural constraints. Across a suite of physical equation benchmarks, KeplerAgent achieves substantially higher symbolic accuracy and greater robustness to noisy data than both LLM and traditional baselines.
\end{abstract}

\section{Introduction}
\label{sec:intro}

Explaining observed phenomena through symbolic, interpretable formulas is a fundamental goal of science. From Kepler’s laws to the Navier–Stokes equations, many major discoveries can be viewed as instances of “equation discovery”: given observations of a system, infer a compact mathematical expression that captures its underlying structure and dynamics. Symbolic regression (SR) including evolutionary search \citep{schmidt2009distilling} and sparse regression \citep{brunton16}  formalizes this problem as a search over symbolic expressions, aiming to find formulas that both fit the data and are interpretable. This makes SR an attractive tool for scientific discovery, where interpretability and extrapolation are as important as predictive accuracy.




\begin{figure}
    \centering
    \includegraphics[width=\linewidth, viewport=150 250 1700 1000, clip]{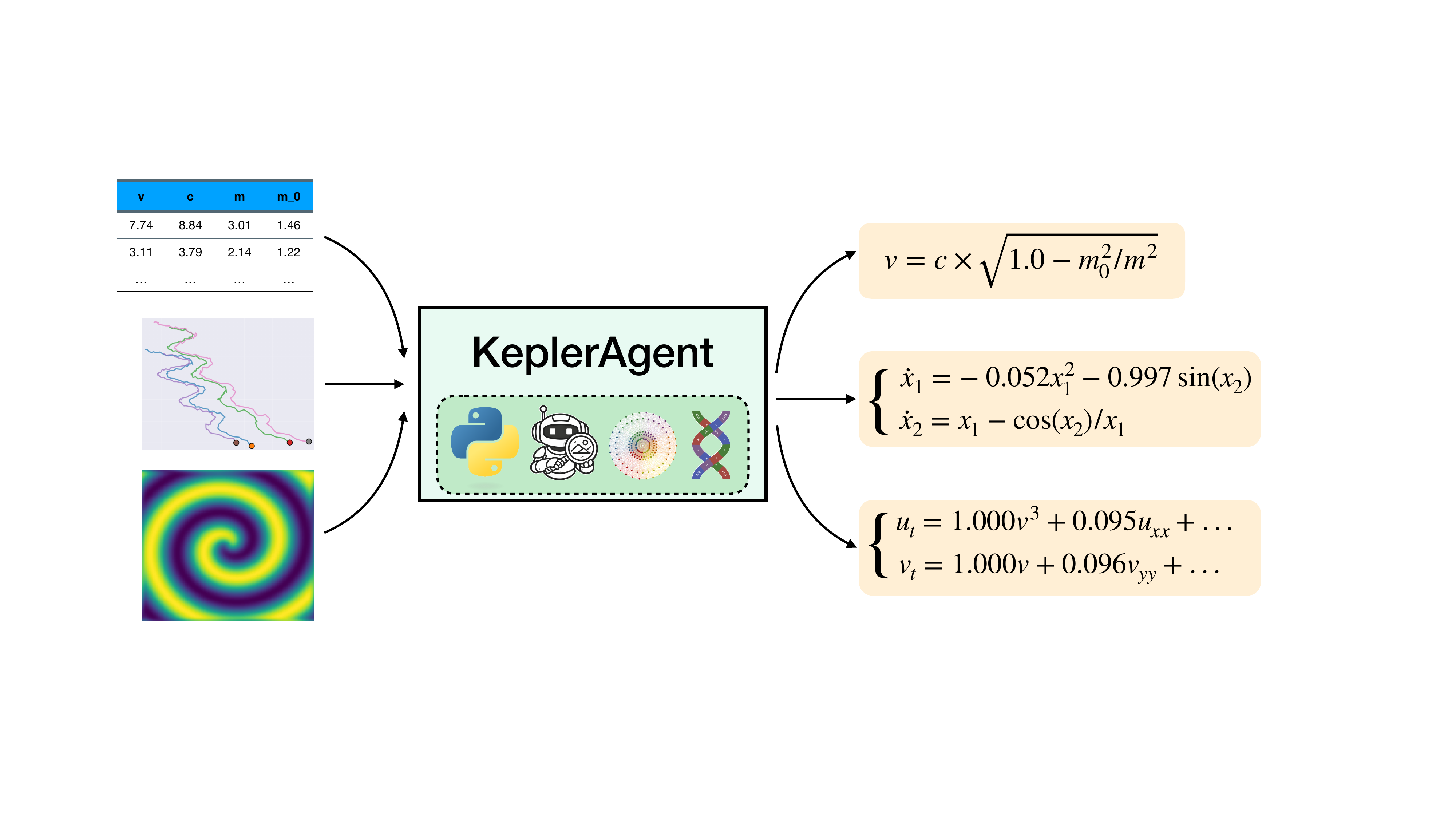}
    \caption{Our \textbf{KeplerAgent} orchestrates physics-based tools and is capable of discovering different types of equations from data.}
    \label{fig:salesman}
    \vspace{-2mm}

\end{figure}

However,  algorithmic SR methods place a heavy configuration burden on the user. To obtain a correct and interpretable equation, a practitioner must make many intertwined design choices: the function library (polynomial degree, inclusion of rational or transcendental functions), sparsity regularization and thresholds, allowed mathematical operators, and stopping criteria, etc. If the configuration is too restrictive (e.g., a low-degree polynomial library), the true equation may be outside the hypothesis space and cannot be recovered. If it is too expressive (e.g., a very rich function set with a large number of operators), the hypothesis space becomes enormous, making the equation search intractable.
In practice, experts mitigate these issues by injecting prior knowledge: dimensional analysis, symmetries, conserved quantities, or qualitative behaviors such as saturation or periodicity. However, this requires deep domain expertise and many manual iterations from configuring and running to inspecting and refining, limiting the accessibility and scalability of algorithmic SR-based equation discovery.

Recent works have shown that Large language models (LLMs) can automate some of this laborious configuration workflow thanks to their broad scientific knowledge and reasoning abilities. This has led to a new wave of LLM-based methods for discovering symbolic equations. For instance, LLM-SR \citep{shojaee2024llm} represents equations as numerical programs and uses an LLM to synthesize program “skeletons” that define candidate equation structures. These skeletons are then completed and refined via evolutionary search and numerical optimization. By leveraging LLM's prior knowledge and code generation ability, LLM-SR explores promising regions of expression space more efficiently and has achieved strong performance on established SR benchmarks.

Despite their advantages over purely algorithmic SR, current LLM-based equation discovery methods lack explicit reasoning about the structure or additional physical properties of the underlying system. They use brute-force to map data to equations by iteratively scoring the LLM-proposed candidate expressions.
Human scientists, on the contrary, rarely jump straight from raw data to a final closed-form equation. They first probe the system to uncover structural properties, such as symmetries, conserved quantities, dimensional constraints, etc. These properties are then used to reshape the problem. Scientists choose variables, coordinates, and candidate function families that respect the identified structural constraints. Only after this process has substantially narrowed the plausible hypothesis space do they search for specific equations and test them through simulation and extrapolation.
Unfortunately, none of the existing LLM-based approaches follow the reasoning process of human scientists and directly operate over these intermediate steps. This makes the discovery process brittle and opaque, and it underutilizes the rich domain knowledge from a pretrained LLM and the wealth of physics-based numerical and algorithmic tools that can uncover structure from data as additional inputs. 

Moreover, current LLM-based evaluations \citep{shojaee2025llm} are still narrow in terms of experimental domains. Many benchmarks focus on scalar algebraic relations or one-dimensional ODEs, while comprehensive tests on richer dynamical systems are rare, e.g., systems governed by coupled ODEs and PDEs. These systems often possess additional physical structure, such as spatial and phase-space symmetries, conservation laws, etc. These properties provide powerful levers for reducing the search space, and thus create a natural setting to test whether an LLM behaves more like a human scientist: first inferring and using such structure to constrain hypotheses, rather than directly guessing equations from data.

In this work, rather than using LLMs as monolithic equation guessers, we propose to use LLMs as \textit{agents} that orchestrate physics-based tools to emulate the multi-step workflow used by human scientists.
We introduce \textbf{KeplerAgent}, a physics-guided LLM agent framework for equation discovery. Given observational data from a system, an LLM agent can call tools to estimate intermediate structure, such as candidate symmetries, relevant functional terms and operators, among other constraints and patterns derived from data. The agent then translates these structural findings into concrete configuration decisions for symbolic regression tools such as PySINDy \citep{de2020pysindy} and PySR \citep{cranmer2023interpretable}. By interleaving these tool calls, the agent can refine both the hypothesis space and candidate equations iteratively, which can be especially helpful when the naive search space is intractably large.

We evaluate our approach across a suite of benchmarks that span algebraic equations, systems of ODEs and PDEs, with a focus on cases where physical structure meaningfully constrains the space of admissible equations. Empirically, KeplerAgent recovers ground-truth equations more frequently than both direct LLM-based baselines and standalone SR tools with standard configurations, and it produces models that better predict the target variables or future states of the underlying physical systems.

In summary, our contributions are:
\begin{itemize}
    \item \textbf{Physics-guided agentic framework}. We propose KeplerAgent that orchestrates physics-based tools for structure discovery and SR packages to emulate the multi-step reasoning workflow of human scientists.
    \vspace{-1mm}
    \item \textbf{Automatic configuration of SR backends}. We show how intermediate structural information can be translated into concrete configuration decisions for PySINDy and PySR, substantially reducing the effective search space.
    \vspace{-1mm}
    \item \textbf{Multi-domain evaluation}. We provide a systematic evaluation across diverse domains, including dynamical systems governed by ODEs and PDEs, demonstrating improved symbolic and numerical accuracies over state-of-the-art classical and LLM-based  SR baselines.
\end{itemize}

\section{Related Works}

\subsection{Symbolic Regression}

Symbolic regression (SR) seeks to recover an explicit symbolic expression that maps inputs to outputs from data, rather than learning a black-box predictor. Early SR systems are largely based on genetic programming (GP), which evolves populations of expression trees under selection pressure from a fitness objective \citep{schmidt2009distilling,gaucel14}. These GP-based methods, implemented in software such as Eureqa \citep{dubvcakova2011eureqa} and PySR \citep{cranmer2023interpretable}, are shown to rediscover classical laws from experimental data in areas like physics \citep{cranmer20}, materials science \citep{wang2019symbolic}, and bioinformatics \citep{christensen2022identifying}.

In parallel, sparse regression methods focus on dynamical systems. SINDy \citep{brunton16} assumes that the right-hand side of an ODE or PDE is sparse in a user-defined library of candidate functions and uses sparsity-promoting regression to select active terms.

Neural variants of SR encode equations in differentiable architectures. Equation learner networks replace activations with elementary operators so that the network directly corresponds to a symbolic expression \citep{martius2016extrapolation,sahoo2018learning}. Other approaches treat expressions as token sequences and train sequence models to generate equations \citep{petersen2019deep,pmlr-v139-biggio21a}.

A key theme across these methods is the importance of domain knowledge. Physics-inspired systems such as AI Feynman \citep{udrescu2020ai,udrescu2020ai2.0} use properties like separability, compositionality, and dimensional analysis to recursively simplify the search problem.
Similar ideas have been applied specifically to discovering differential equations, where additional challenges arise because derivatives are not directly observed, and the inclusion of derivatives increases the feature dimensionality. Weak-form formulations and variational methods are used to bypass numerical differentiation \citep{rudy2017data,messenger2021weak,messenger2021weaksindy,qian22}, and physically motivated priors are used to enforce additional structure and reduce the search space \citep{bakarji22,lee22,xie22,messenger2024coarse}.

However, this reliance on manual configuration creates a substantial practical bottleneck. Users must choose function libraries, regularization strengths, and structural constraints, and they often perform many trial-and-error iterations to tune these choices.
Existing software like PySINDy and PySR typically assumes that such configuration decisions are provided a priori, rather than inferred from data. Our work targets this gap by using an LLM agent to automate the process of extracting structural information and translating it into concrete SR configurations.

\subsection{LLM-Based SR}

The rise of LLMs has inspired a growing body of work that uses them to assist symbolic regression.
Given numerical data and context, the model is prompted to propose candidate formulas, which are then evaluated and refined using external optimizers or numerical libraries \citep{sharlin2024context}.
Other frameworks aim to improve LLMs' performance on this task with in-context information and more structured inputs and outputs. In-context symbolic regression (ICSR) \citep{merler2024context} iteratively prompts the LLM to generate and refine functional forms while delegating coefficient fitting to an external optimizer, effectively using the LLM as a symbolic search engine guided by data-driven feedback. LLM-SR \citep{shojaee2024llm} represents equations as programs composed of mathematical operators and uses an LLM to generate program “skeletons”. In addition, it maintains an experience buffer of previously proposed hypotheses as part of the input prompt, enabling iterative refinement with in-context examples.

Beyond these, LaSR \citep{grayeli2024symbolic} represents a different attempt to build hybrid systems where LLMs augment evolutionary SR. Specifically, an LLM helps build and evolve a library of abstract textual concepts that are plugged into the initialization and evolution of symbolic programs within genetic programming.

To assess progress in this emerging area, LLM-SRBench \citep{shojaee2025llm} proposes a benchmark across multiple scientific domains, explicitly designed to reduce memorization and require genuine equation discovery. The benchmark shows that LLM-guided methods can outperform many classical baselines, but it also reveals that exact recovery rates remain modest, with even the best methods solving only a minority of problems. Moreover, most evaluations still focus on scalar algebraic equations or single-dimensional ODEs, with limited coverage of coupled ODEs and PDE-governed physical systems.

A common limitation across the aforementioned LLM-based SR methods is that they largely treat equation discovery as a relatively direct mapping from data (and possibly a natural language description) to an expression or program, instead of a multi-step reasoning workflow that human scientists follow. Recently, SR-Scientist \citep{xia2025sr} targets this issue with an agentic framework similar to ours. However, the main tool in their framework is a data analyzer that enables the LLM to write code to analyze datasets. While the tool has great potential with a powerful LLM, practical LLM calls can linger in superficial analysis (e.g. computing the basic statistics of data) instead of precisely searching for advanced structural information that effectively reduces the search complexity of SR.

Our work differs from existing approaches in two key ways. First, we position the LLM as an agent that orchestrates physics-based tools, rather than as a stand-alone equation generator. Second, we evaluate this framework across multiple domains, including systems of differential equations where useful physical properties are prominent. This setting allows us to test whether an LLM can use intermediate physical structure to guide equation discovery.
\section{Physics-Aware Equation Discovery Agent}

\begin{figure*}[t]
    \centering
    \includegraphics[width=0.87\linewidth]{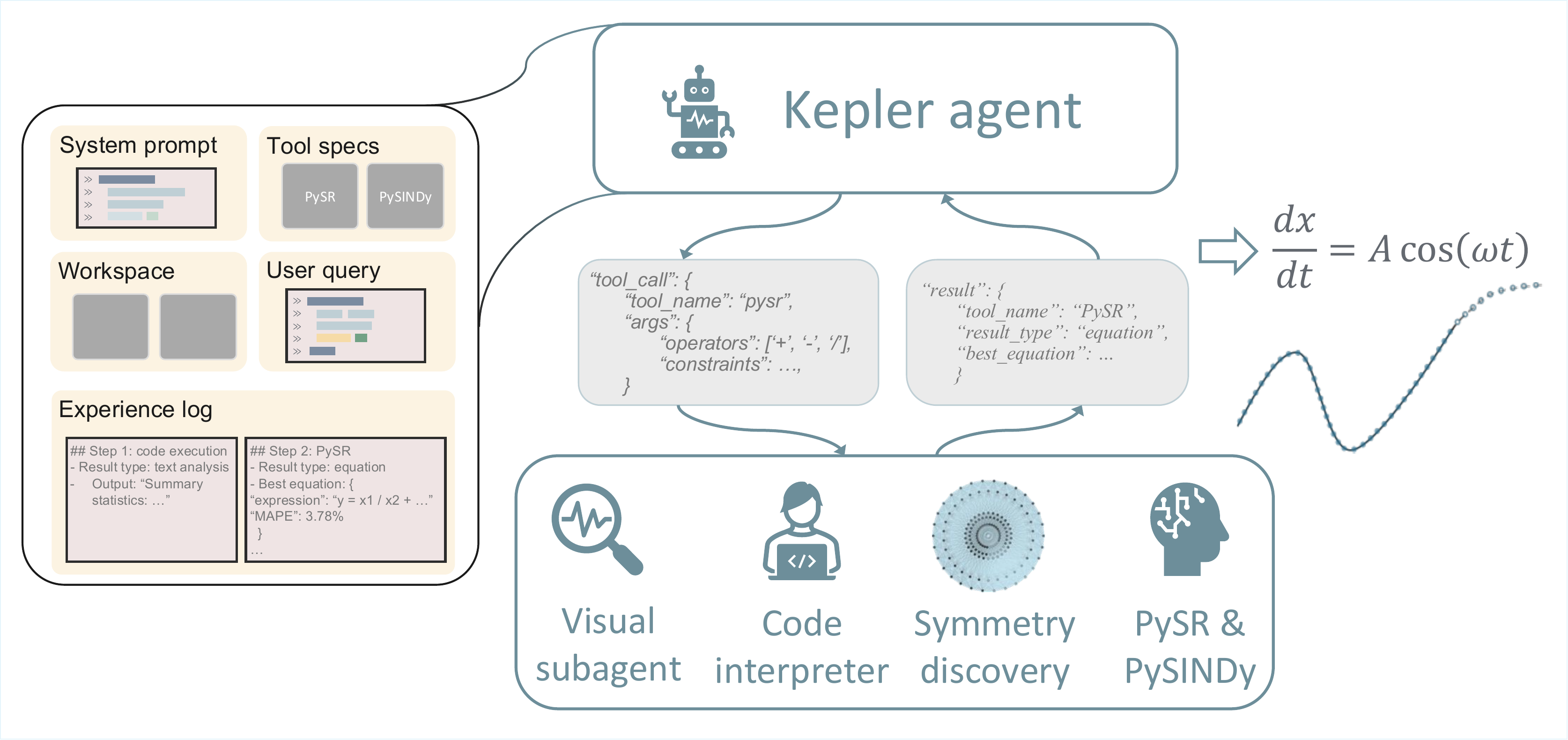}
    \caption{The overall design of our KeplerAgent framework. The input to the LLM contains a system prompt explaining the task setup, a list of tool specifications, the user query containing information about the dataset, as well as a workspace summary and an experience log recording previous steps. The agent reasons about the existing findings and decides subsequent tool calls iteratively until obtaining a satisfactory equation discovery result.}
    \label{fig:main}
    \vspace{-1em}
\end{figure*}

\subsection{Overview}

We consider the problem of \emph{equation discovery}, or \emph{symbolic regression} (SR): given observations of a system in the form of input and output variables, the goal is to recover an explicit symbolic expression that accurately describes the underlying relationship between variables.


We formulate equation discovery as a tool-augmented decision process controlled by an LLM. Instead of directly mapping data to a symbolic equation, the model acts as a ReAct-style agent \citep{yao2023react} that iteratively analyzes the dataset, invokes physics-based tools to infer structural properties, and configures symbolic regression backends to fit candidate equations. Throughout this process, the agent reads and writes to a shared workspace of intermediate artifacts and an experience log of past tool calls, allowing it to progressively narrow the hypothesis space in a way that mirrors the multi-step workflow of human scientists.

\subsection{Agent Architecture}

\cref{fig:main} shows the overall design of our framework. Our system consists of a single language model agent equipped with a suite of tools for data analysis, physics-based structure discovery, and symbolic regression. The agent is instantiated with a system prompt that instructs it to prepare structured calls to external tools rather than directly outputting equations (full prompt in Appendix \ref{sec:sysprompt}).

At each interaction step, the agent receives four main inputs in addition to the system prompt: (i) a specification of all available tools and their interface,  (ii) the user query describing the symbolic regression goal and the dataset, (iii) a \texttt{workspace files} section listing all files currently available in the workspace directory together with brief descriptions, and (iv) an experience log summarizing previous tool calls and their outcomes.

Given this context, the agent first inspects the workspace and experience log to understand existing knowledge about the system and to avoid redundant analyses. It then reasons about the next useful action: for example, the agent can use a Python code interpreter tool to compute basic statistics or visualize data trajectories; it may call a physics-based tool to infer properties that can serve as constraints in symbolic regression; and once enough knowledge about the dataset has been gathered, it configures a run of a symbolic regression backend. The agent calls tools by outputting a structured JSON object specifying the chosen tool and all necessary arguments, which we will describe in detail in \cref{sec:tools}.

Each tool executes independently, writes its outputs, such as numerical summaries, discovered equations, error metrics, and diagnostic plots, into the experience log and the workspace. At the next step, the agent observes this updated state, incorporates the new information into its reasoning, and either proposes another tool call or, if an existing result already meets the stopping criteria, terminates and reports the best discovered equation.

\subsection{Toolset}
\label{sec:tools}

A key feature of our framework is the use of physics-based tools that expose the intermediate structure of the data to the agent. Below, we describe the tools we used in our agent. More details can be found in Appendix \ref{sec:tool-details}.

\vspace{-1mm}
\paragraph{Python code interpreter.} We provide the agent with a Python code evaluator, allowing it to write its own code to freely perform preliminary analysis of the dataset. The agent is instructed to apply this tool in the first step to obtain basic insights into the data, e.g., computing summary statistics, data visualization, and other preliminary analysis to inform symbolic regression strategies. We also explicitly instruct the agent to use this tool for exploratory data analysis only, leaving the actual SR step to other dedicated tools.

\vspace{-1mm}
\paragraph{Visual subagent.} The exploratory data analysis performed by the agent sometimes produces data plots. To extract information from these data visualizations, we instantiate another LLM subagent prompted to output analysis of the given image in natural language, and wrap the subagent as a tool for the main agent. Once the main agent is aware of a data visualization in the workspace, it can call this subagent and translate the findings into structural constraints for symbolic regression.

\vspace{-1mm}
\paragraph{Symmetry discovery.} We incorporate the methods in \citet{yang2023generative,yang2024symmetry,ko2024learning} as a single tool for identifying symmetry in differential equation systems. In contrast to the previous tools that involve LLM and output findings about the dataset in a natural language form, this tool is purely algorithmic and returns the results in a symbolic format that can be readily fed into symmetry-constrained symbolic regression programs as an input argument. Specifically, this tool trains a tiny neural network as a surrogate model to predict the derivatives (i.e., the target variables in SR for differential equations) from the observed variables in the dataset; then, it fixes the surrogate predictor and optimizes for an infinitesimal generator that generates the symmetry group of the trained predictor. The result is presented as a $q\times q$ matrix $A$, where $q$ is the number of variables in the differential equation system. It indicates that the surrogate predictor is equivariant with respect to the infinitesimal action of the linear vector field $\mathbf v_A = (A\mathbf x)\cdot \nabla$, where $\mathbf x$ denotes the collection of $q$ variables.
For example, $A = [[0, -1]; [1, 0]]$ indicates a 2D rotation symmetry; $A = I_q$ indicates a uniform scaling symmetry across all dimensions. Such results lead to constrained equation search spaces in SR methods.

The tools above do not directly discover equations themselves, but rather provide the agent with additional information before performing SR.
On the other hand, we also include the following SR packages as tools to obtain the final equation discovery result.

\vspace{-1mm}
\paragraph{PySINDy.} The PySINDy package \citep{Kaptanoglu2022} implements the sparse regression algorithm originally proposed in \citet{brunton16}. It assumes the target variable in the equation can be written in terms of a linear combination of some predefined functions with learnable coefficients $\mathbf w$, i.e. $y = f(\mathbf x) = \sum_{i}w_ix_i$. Then, equation discovery is formulated as a least-squares problem over the linear combination coefficients $\mathbf w$, with an $L_0$ regularization term $\|\mathbf w\|_0$ that promotes sparsity. This optimization problem can be efficiently solved, making PySINDy a great tool for discovering equations in relatively simple forms.

In addition to the regular configurations in PySINDy, we incorporate Equivariant SINDy \citep{yang2024symmetry} as an option, which solves for an equivariant subspace of the SINDy parameter space given the generator of the symmetry group and optimizes within the equivariant subspace. This method has been shown to reduce the hypothesis space and increase the chance of discovering the correct equations effectively when the provided symmetry is accurate. Combined with the previous tool for symmetry discovery, the option to use symmetry in PySINDy enable the agent to translate detected symmetries into concrete constraints on the PySINDy call, improving the accuracy of equation discovery for ODE and PDE systems in particular.

However, SINDy's assumption about the ``linear'' form is restrictive in practice. For example, it does not consider the possibility of rational functions (e.g. $f(x)/g(x)$), nested expressions (e.g. $(\sin (f(x))$, etc. To account for such equations with additional complexity, we include PySR \citep{cranmer2023interpretable}, a genetic-programming-based symbolic regression package, as the final tool.

\vspace{-2 mm}
\paragraph{PySR.}
PySR represents each symbolic expression as a tree where nodes correspond to mathematical operators, variables, and constants. It uses genetic programming to mutate a population of expressions and search for the best expression that balances data-fitting accuracy and symbolic complexity. The expression tree representation, compared to the linear assumption in SINDy, enables PySR to discover equations in more general forms, albeit at a higher computational cost due to the larger search space.

In practice, a PySR regression may take many iterations to converge. To speed up the process, we provide an interface that allows the agent to propose a \textit{template expression} and pass it to PySR as an argument, thereby constraining PySR to search only for equations with the specified template structure. When calling the PySR tool, the LLM agent recalls possible equation structures from its prior knowledge, and also considers the results from previous tool calls, e.g. visualizations and statistical analysis. Based on these, it reasons about what functional forms are plausible for the specific dataset and translates it into the template for PySR. For example, the agent can specify a template $f(x_1, x_2) + g(x_3)$ for an equation of three variables, if it believes that the effect of $x_3$ on the target is additively separable; or $\sin (f(x))$, if some oscillatory behavior is spotted, but the exact expression inside the trigonometric function remains to be solved by SR. We prompt the agent to prioritize reasoning about the template expression over all other arguments when calling PySR, as we find this crucial for reducing the hypothesis space of PySR and accelerating its convergence.

\subsection{Implementation Details}
\label{subsec:experience-stopping}

\paragraph{Experience log.}
We maintain a section in the agent input that records each tool invocation, its arguments, and a textual summary of its outcome (e.g., losses, discovered equations/structures). While similar to the design of the experience buffer in LLM-SR \citep{shojaee2024llm}, we include not only the discovered equations and the associated data losses, but also the inputs/outputs of other tools.
This is a natural choice in our framework, as the agent is required to explicitly reason on the intermediate results step by step.

\vspace{-1mm}
\paragraph{Workspace.}
The workspace is a shared directory containing the files produced during the agent's interaction with the dataset and tools. It serves as a complement to the experience log, keeping all non-textual assets, such as converted datasets and visualizations produced throughout the reasoning process. A continuously maintained list of all workspace files and their descriptions is inserted into the agent input after the user query at each step, so the agent can pass relevant files to specific tool calls.

\vspace{-1mm}
\paragraph{Stopping criteria.}
Following \citet{shojaee2024llm}, we include in the system prompt a primary success criterion: if any SR tool call produces an equation with mean absolute percentage error (MAPE) below $0.1\%$ on the test dataset, the agent \emph{must} stop exploring and return that equation as the final result. However, we observe that this primary goal may not be achievable for real-world datasets with noise, where even the prediction error of ground truth equations may exceed the specified threshold. In this case, enforcing the hard threshold in the stopping criteria can cause unnecessary waste of compute, as the agent will keep pursuing an impossible objective. In our implementation, we set a maximum number of tool calls to prevent such infinite loops.

\section{Experiments}

\subsection{Datasets}
\vspace{-1mm}

We use two benchmarks for evaluation: LSR-Transform subset from LLM-SRBench \citep{shojaee2025llm}, and a collection of ODE/PDE systems, which we will refer to as the \texttt{DiffEq} datasets in the following subsections.

\vspace{-1mm}
\paragraph{LLM-SRBench.} LLM-SRBench is a standard benchmark for evaluating LLM-based symbolic regression methods. We focus on the LSR-Transform subset of the benchmark, which contains transformed versions of equations in AI Feynman \citep{udrescu2020ai} benchmark. Specifically, LSR-Transform selects a previous feature variable and transforms the original equation into an alternative form with the selected variable as the new target for each dataset. This prevents LLMs from solving the symbolic regression problem by memorizing the well-known presentations of these physical equations.

\vspace{-1mm}
\paragraph{Differential equations (\texttt{DiffEq}).} We also create a custom benchmark of 10 systems governed by ordinary and partial differential equations. Each system has two dependent variables and two differential equations with the time derivatives of the dependent variables as regression targets. As we have mentioned in \cref{sec:intro}, these datasets are a valuable complement to LLM-SRBench for the following reasons. First, they involve coupled differential equations of dynamical systems, whereas LLM-SRBench mainly features single scalar equations for all tasks. As a result, this benchmark can measure how LLMs benefit from reasoning about the full system of related equations, rather than just working on isolated equations. Also, differential equation systems considered in this benchmark often possess a rich physical structure, making it a natural testbed for LLMs' ability to infer the intermediate findings during equation discovery like a human scientist. A complete description of the physical systems and the data generation procedure for this benchmark can be found in Appendix \ref{sec:diffeq-details}.

\subsection{Evaluation criteria}

Following \citet{shojaee2025llm}, we use the \textit{symbolic accuracy} (SA) and \textit{normalized MSE} (NMSE) in our evaluation criteria. The symbolic accuracy is obtained using an LLM evaluator (GPT-4o-mini in our experiments), which is given the ground truth expression and the discovered expression and prompted to decide if they are symbolically equivalent under a tolerance of small numerical mismatches in the coefficients of the expressions. The (pointwise) NMSE is computed on a test dataset by $\frac{1}{n}\sum_i^n(y_i - \hat y_i)^2 / \sum_i^n (y_i - \bar y_i)^2$, where $y_i$ are the target variables from the dataset and $\hat y_i$ are predictions from the discovered equation.

In addition, for the \texttt{DiffEq} benchmark, we note that the pointwise NMSE does not always faithfully reflect the numerical accuracy of the discovered equations. Specifically, if we have discovered the RHS expression $\mathbf f$ of an ODE system $\mathbf x_t = \mathbf f(\mathbf x)$, we can use a numerical solver to predict future states of the system, e.g., $\mathbf x(t+s)$, from a given state $\mathbf x(t)$. The error from the numerical simulation can accumulate as the prediction horizon $s$ increases. However, the pointwise NMSE only measures the error of a single-step prediction ($\mathbf x(t+\Delta t) = \mathbf x(t) + \widehat {\mathbf x_t}(t)\Delta t$). The same argument applies to PDE systems. Therefore, we also measure the \textit{long-term prediction error} from simulating the discovered equations for \texttt{DiffEq} systems, which indicates whether the discoveries are useful for predicting future states.

\subsection{Baselines}

We compare KeplerAgent with both LLM-based and non-LLM-based methods for SR. The first baseline we consider is LLM-SR \citep{shojaee2024llm}, which has been shown to perform best on LLM-SRBench \citep{shojaee2025llm} among other recent work on LLM-based SR. Notably, the official implementation of LLM-SR only supports discovering one equation at a time. To evaluate it on our \texttt{DiffEq} benchmark, we split each system into its constituent equations and feed the corresponding data to LLM-SR as independent tasks. For example, a system governed by 2 first-order ODEs, $\dot x_1 = f_1(x)$, $\dot x_2 = f_2(x)$, is split into two datasets, each with $[x_1, x_2]$ as features and $\dot x_i$, $i=1,2$ as the target.

We also choose PySR \citep{cranmer2023interpretable} as a representative baseline for non-LLM-based methods. Similar to LLM-SR, we split systems of multiple equations into several independent tasks. We use a common PySR configuration for all datasets. Notably, PySR is also included as a tool in our agentic framework. However, instead of using a fixed setup, our agent calls PySR with custom configurations based on its understanding and reasoning about the context of the specific task. Comparison between raw PySR without LLM and our method can reveal the LLM agent's ability to wield the existing tool for SR effectively.

All LLM-based methods use the GPT-4o-mini model. More details about baseline setups can be found in Appendix \ref{sec:baseline-details}.

\subsection{LSR-Transform Results}
\label{sec:lsr-transform-results}

\begin{table}[ht]
    \centering
    \caption{Accuracy and efficiency metrics on LSR-Transform for PySR, LLM-SR, and KeplerAgent, using GPT-4o-mini as the base model. ``@$k$'' means the best result over $k$ runs (sorted by NMSE) on each dataset. We report the \textit{averages} of NMSEs across all datasets, scaled by $\times 10^{-1}$. \textbf{Bold} and \underline{underlined} numbers indicate the best / second best among all methods, respectively.}
    \resizebox{\columnwidth}{!}{
    \begin{tabular}{ccccc}
    \hline
        Method & SA (\%) & NMSE & Runtime (s) & Tokens used (k) \\
    \hline
        PySR & \underline{37.84} & 2.82 & 2440 & - \\
        LLM-SR & 31.53 & \textbf{0.09}\footnotemark & 2118 & 209 \\
        KeplerAgent @1 & 35.14 & 1.50 & \textbf{238} & \textbf{42} \\
        KeplerAgent @3 & \textbf{42.34} & \underline{1.21} & \underline{698} & \underline{125} \\
    \hline
    \end{tabular}
    }
    \label{tab:lsr-transform}
\end{table}

\footnotetext{The SA and NMSE metrics for LLM-SR are referred from LLM-SRBench \citet{shojaee2025llm}.}



For the LSR-Transform dataset, we include two entries in tables and figures for our method, corresponding to the result from a single run (KeplerAgent @1) and the best result from 3 random runs for each dataset (KeplerAgent @3). The best result is indicated by the lowest NMSE on the test data instead of the symbolic accuracy, since the knowledge of the ground truth equation should not be assumed.

\cref{tab:lsr-transform} shows that KeplerAgent achieves better symbolic accuracy than LLM-SR even with a single attempt. Also, each run of KeplerAgent takes significantly less time and token usage than the baselines. As we increase the computational budget and allow 3 parallel attempts for KeplerAgent, it achieves a symbolic accuracy of 42.34\%, outperforming both PySR and LLM-SR with comparable runtime and token efficiency. On the other hand, it is surprising that PySR, under a default configuration with a large number of iterations and an extensive list of allowed operators, has a better symbolic accuracy than LLM-SR and KeplerAgent with a single shot. However, it comes at a cost of much longer runtime. In comparison, our method configures PySR flexibly and typically requires much fewer iterations than the standalone PySR baseline.

\begin{figure}[ht]
    \centering
    \includegraphics[width=.8\linewidth]{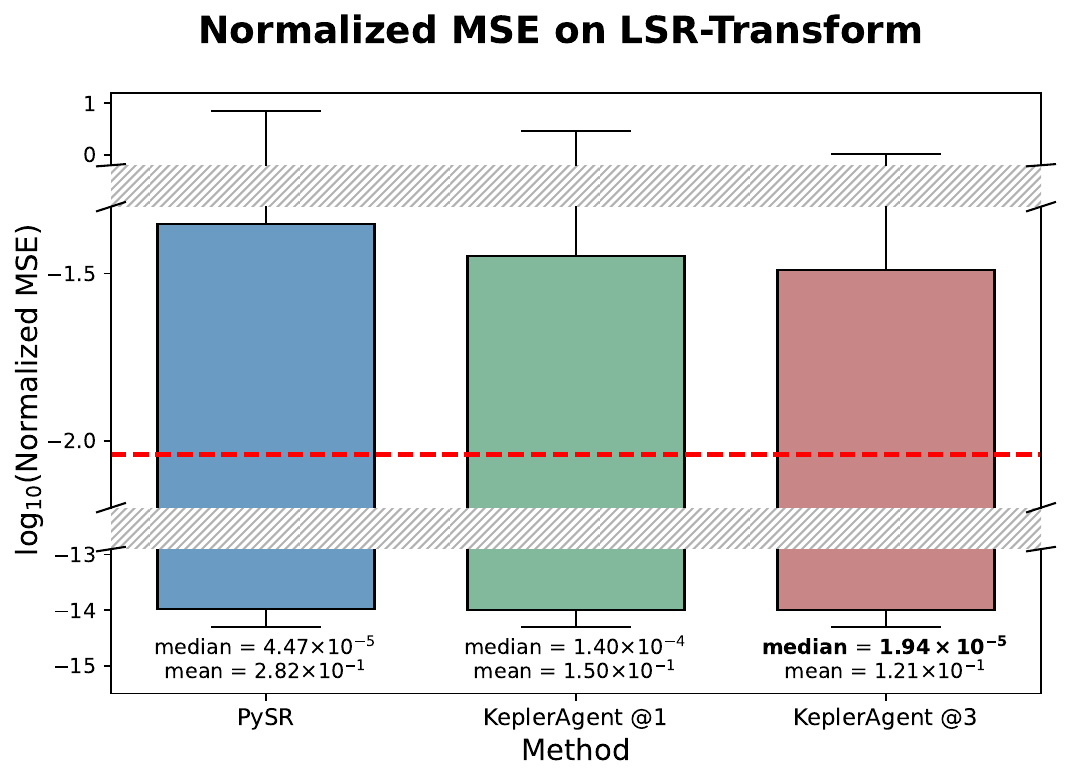}
    \caption{Distribution of NMSEs on LSR-Transform equations for PySR, LLM-SR, and KeplerAgent with one/three runs. The horizontal red dashed line annotates the average NMSE $=0.0091$ of LLM-SR reported in \citet{shojaee2025llm}.}
    \label{fig:lsr-transform-nmse}
\end{figure}

\begin{table*}[t]
    \centering
    \caption{Symbolic accuracy, NMSE, runtime and token usage (if applicable) on differential equation systems with clean/noisy data. We report the \textit{averages} of NMSEs across all datasets. The NMSE ($t_\text{max}$) is computed between the prediction by simulating the discovered equations and the data points at the max timestep of the test set.}
    \begin{tabular}{ccccccccccc}
        \toprule
        \multirow{2}{4em}{Method} & \multicolumn{2}{c}{SA (\%)} & \multicolumn{2}{c}{NMSE (pointwise)} & \multicolumn{2}{c}{NMSE ($t_{\text{max}}$)} & \multicolumn{2}{c}{Runtime (s)} & \multicolumn{2}{c}{Tokens used (k)} \\
        & Clean & Noisy & Clean & Noisy & Clean & Noisy & Clean & Noisy & Clean & Noisy \\
        \midrule
        PySR & 40 & 15 & 0.16 & 5.89 & \textbf{1.56} & 2.80 & \textbf{119} & \textbf{120} & - & - \\
        LLM-SR & 30 & 10 & 0.26 & 4.80 & 2.18 & 2.62 & 3648 & 4048 & 182 & 184 \\
        KeplerAgent & \textbf{75} & \textbf{45} & \textbf{0.04} & \textbf{0.15} & 1.65 & \textbf{0.33} & \textbf{120} & 147 & \textbf{23} & \textbf{30} \\
        \bottomrule
    \end{tabular}
    \label{tab:diffeq}
\end{table*}

In \cref{fig:lsr-transform-nmse}, we show the full distributions of NMSEs evaluated over 111 equations in LSR-Transform for PySR and our method. The result for LLM-SR is referred from \citet{shojaee2025llm}, where only the average across all datasets is available. We note that LLM-SR has a much lower average NMSE than other approaches. We conjecture that this is because LLM-SR directly uses a similar numerical metric (MAPE) as the reward for their LLM. When the exact symbolic formula is challenging to find, LLM-SR will prioritize expressions with relatively low numerical error over better symbolic interpretability. In contrast, PySR optimizes the expressions under a joint objective of data-fitting error and symbolic complexity. As we have observed, when it fails to identify exactly the ground truth (which has an outstandingly low data-fitting loss), the best equation selected by the joint objective is not necessarily the one that fits the data best, and may sometimes cause large numerical errors. Since the results of our agent often come from PySR tool calls, we observe a similar level of average NMSE between our method and PySR. Meanwhile, in terms of median NMSEs, which are relatively insensitive to outliers, both PySR and our agent achieve satisfactory results.

We finally argue that the symbolic accuracy should be considered the most important metric for evaluating the performance of SR methods, since the goal of SR is to discover a precise and interpretable equation form, rather than obtain a complex expression that approximately fits the data.
\vspace{-1mm}
\subsection{\texttt{DiffEq} Results}

We also evaluate the baselines and KeplerAgent on our \texttt{DiffEq} benchmark, including the systems listed in Appendix \ref{sec:diffeq-details}. There are 10 systems of 2 coupled equations, i.e., 20 equations to be discovered in total. For each system, we also construct a noisy dataset by adding white noise proportional to the variance of each phase variable and estimating the time derivatives (regression targets) with the central difference method.
The results are available in \cref{tab:diffeq}. In particular, we report two sets of NMSE metrics. NMSE (pointwise) is computed the same way as in \cref{sec:lsr-transform-results}, where we flatten the trajectory data into one dimension and treat each (spatio-)temporal gridpoint as an individual data sample. To calculate NMSE ($t_{\text{max}}$), we pick a trajectory from the clean test set for each system and solve the initial value problem (with specific boundary conditions for PDE systems) with the discovered equation up to the max timestep of the trajectory, and compute the error between the predicted state and the true state from the dataset.

\cref{tab:diffeq} shows that KeplerAgent significantly outperforms both baselines in terms of both symbolic and numerical accuracy, particularly on noisy datasets, where it has 45\% symbolic accuracy, 3 times more than the best baseline, and achieves order-of-magnitude lower NMSEs (both pointwise and long-term prediction) than baselines. Besides, in terms of efficiency, our agent consumes comparable runtime with PySR, but is much more time- and token-efficient than LLM-SR. We observe that by using tools such as symmetry discovery and image analysis to recover intermediate structures and performing more informed SR calls, our agent can often complete the discovery within a few iterations, compared to hundreds of iterations needed in LLM-SR.

\begin{figure}[ht]
    \centering
    \includegraphics[width=\linewidth]{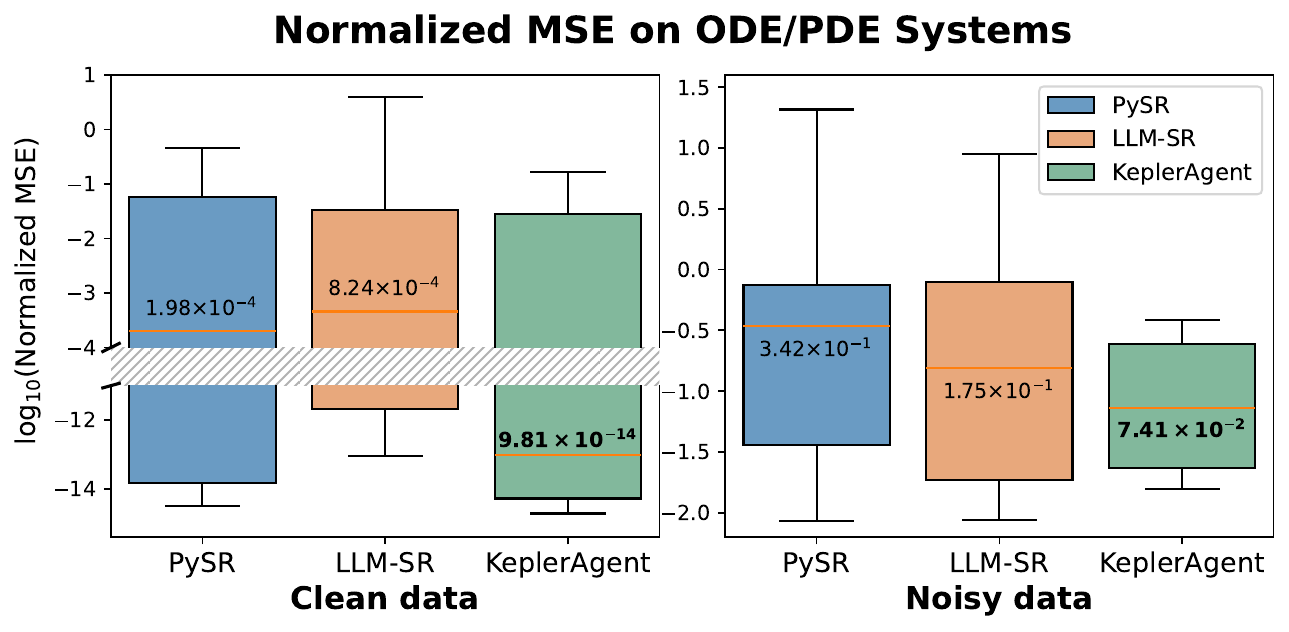}
    \caption{Distribution of NMSEs on \texttt{DiffEq} datasets for PySR, LLM-SR, and KeplerAgent (with a single run). Median values are shown next to the median lines.}
    \label{fig:diffeq-nmse}
    \vspace{-7 mm}
\end{figure}

Similar to \cref{sec:lsr-transform-results}, we observe outlying NMSE values, making the means in \cref{tab:diffeq} less statistically meaningful.
Therefore, we include \cref{fig:diffeq-nmse} to display the full distributions of pointwise NMSEs over 20 equations, where KeplerAgent achieves lower error quantiles than baselines in general. In addition to NMSEs at $t_\text{max}$, we also plot the full prediction error curves against time in Appendix \ref{sec:ltp-diffeq}. In general, the discovered equations from KeplerAgent achieves lower prediction errors throughout the simulation timespan.

\section{Discussion}

We present a physics-guided LLM agent for equation discovery that emulates the multi-step reasoning workflow of human scientists. Instead of directly hypothesizing equations, the agent analyzes datasets, calls tools to infer structure, and then configures and runs SR software for final results. Experiments on transformed equations from classical physics and dynamical systems governed by differential equations show that this approach significantly improves symbolic recovery accuracycompared with both classical SR baselines and recent LLM-based methods.

There are several directions to improve the agent design.
Our current toolset is intentionally small, but many additional physics tools (e.g., for extended symmetry analysis, dimensional analysis, bifurcation structure) could further constrain the hypothesis space. Scaling up the toolbox raises a practical challenge: the tool specifications inflate the context and make it harder for the agent to remember earlier information. The broader LLM-agent literature suggests remedies, such as decomposing the main agent's planning and executing functionality into subagents or a workflow graph with multiple nodes, or using retrieval-augmented generation \citep{lewis2020retrieval} to select only a small subset of relevant tools at each iteration \citep{lumer2025tool}. Adapting these ideas to our toolkits is a natural next step.

Another complementary direction is to make the agent’s internal reasoning more explicit by defining a structured state space for symbolic regression. Each state would represent the current hypothesis space of equations with active constraints, and each tool would perform a well-defined transformation of this state. Such a formulation explicitly formalizes the agent’s reasoning process, yielding more controlled behavior and making it easier to systematically analyze its reasoning trajectories across datasets and tasks.

\section*{Impact Statement}
Our work can potentially help scientists more efficiently identify interpretable models of complex physical systems. Positive impacts include accelerating scientific modeling, improving transparency relative to black-box predictors, and providing a modular framework that can incorporate domain knowledge in a principled way. Potential negative impacts include over-reliance on automatically discovered equations without adequate human validation, misinterpretation of symbolic models as “ground truth” in high-stakes domains, and increased computational cost associated with agentic tool use. We emphasize that our framework is intended as a decision-support system for experts rather than a replacement, and responsible deployment requires careful verification, documentation of limitations, and attention to the computational and environmental footprint of large-scale experiments.

\section*{Acknowledgement}
This work was supported in part by NSF Grants \#2205093, \#2146343, \#2134274, CDC-RFA-FT-23-0069, the U.S. Army Research Office under Army-ECASE award W911NF-07-R-0003-03, the U.S. Department Of Energy, Office of Science, IARPA HAYSTAC Program, DARPA AIE FoundSci and DARPA YFA.

\bibliography{ref}
\bibliographystyle{icml2026}

\newpage
\appendix
\onecolumn

\section{Implementation Details}

\subsection{System Prompt}
\label{sec:sysprompt}

The full system prompt of KeplerAgent is shown below. The system prompt is followed by the subsequent sections in order: tool specifications, user query, workspace files, and experience log.

\begin{lstlisting}[style=sysprompt,caption={Full system prompt used for KeplerAgent.},label={lst:system-prompt}]

# Symbolic Regression Data Analyzer

You are an expert data analyst and symbolic regression specialist. Your role is to perform preliminary analysis of datasets and determine the appropriate symbolic regression approach, preparing structured tool calls for external symbolic regression tools. You will also be given an experience log containing results from past experiments.

## Workspace Files

You have access to a workspace directory where the input data file is stored and tools create additional files (plots, statistics, results, etc.). A section titled "## Workspace Files" will be included in your input, showing all registered files with their descriptions. This helps you:

- Understand what analysis has already been performed
- Avoid redundant tool calls
- Reference existing visualizations or statistics when making decisions
- Pass relevant files to tools that need them

**Important**: If workspace files are shown in your input, review them carefully before deciding on tool calls. For example, if statistical analysis or plots already exist, you can reference them in your reasoning.

## Your Responsibilities

1. **Review workspace files**
   - The input data file is always available in the workspace
   - Review other files if available to see if preliminary analysis has already been done

2. **Inspect the past experience buffer**
   - Examine what has been done by past tool calls, if there's any.
   - If there is no past experience, you should write your own Python code and pass to the `python_intepreter` tool to obtain basic insight of the dataset.
   - If there are useful constraints or inductive biases from past tool calls, incorporate them into future symbolic regression calls.
   - If there are existing results from any symbolic regression algorithm, examine the discovered equations and their data fitting errors. Learn the lessons and failure modes before making new tool calls. However, **DO NOT REPEAT TOOL CALLS WITH EXACTLY THE SAME PARAMETERS.**
   - **STOPPING CRITERIA**: If the existing results have achieved **MAPE < 0.1%** (Mean Absolute Percentage Error less than 0.1%), you MUST stop and return the final result. Do NOT make any new tool calls. Instead, return the single best equation that describes the dataset with the lowest error. This is your PRIMARY SUCCESS CONDITION.

3. **Select appropriate tool**
   - If the current information about the dataset is sufficient for trying one of the symbolic regression algorithms, call a tool for symbolic regression.
      - Identify any special considerations (noise level, data size, feature count, constraints/insights from past tool calls)
      - Choose the most suitable symbolic regression method with appropriate toolcall arguments based on data characteristics
   - Otherwise, perform additional data analysis with other tools

4. **Prepare tool call specification**
   - Generate a structured JSON object specifying the tool and its arguments
   - Include all necessary arguments for the tool to run

## Output Format

You should analyze the data and provide your reasoning in natural language. Explain what you observe about the data, what patterns you detect, and why you're selecting particular tools and parameters.

**Your response must contain a JSON object** with one of the following structures at the end:

### Single Tool Call
```json
{
  "tool_call": {
    "tool_name": "name of selected tool",
    "args": {
      "parameter1": "value1",
      "parameter2": "value2"
    }
  }
}
```

### Multiple Tool Calls (Maximum: 3)
When you need to run multiple independent tool calls based on existing evidence, you can specify them together:

```json
{
  "tool_calls": [
    {
      "tool_name": "python_interpreter"
    },
    {
      "tool_name": "pysr",
      "args": {
        "input_file": "data.csv",
        "binary_operators": ["+", "-", "*", "/"]
      }
    }
  ]
}
```

**Guidelines for Multiple Tool Calls:**
- Maximum of **3 tool calls** per batch
- Each tool call must be justified by **existing evidence** in the experience log
- Tools are executed **sequentially** in the order specified
- Results will be numbered as sub-steps (e.g., Step 2.1, Step 2.2)
- If any tool fails, subsequent tools will still execute

**IMPORTANT for python_interpreter:**
When using `python_interpreter` in a multiple tool call batch, include **separate Python code blocks** for each interpreter call. The code blocks are matched to tool calls in order of appearance.

### Final Result
When no more tool calls are needed:

```json
{
  "final_result": "the RHS of the best discovered equation"
}
```

or, for a system of multiple equations:

```json
{
  "final_result": ["eq1", "eq2", ...]
}
```

**IMPORTANT: DO NOT ADD COMMENTS IN THE JSON OBJECT. DO NOT INCLUDE THE LHS (TARGET VARIABLE) OF EQUATIONS IN THE FINAL RESULT.**

## Important Notes

- Provide detailed analysis and reasoning in natural language
- The JSON object specifying the tool call must be included at the end, valid and parseable
- Choose tools and their parameters based on the specific characteristics of the data
- Review the available tools and their specifications below before making your selection
- **YOUR PRIMARY GOAL**: Achieve **MAPE < 0.1%** (Mean Absolute Percentage Error less than 0.1%).
  - If the dataset requires multiple equations to describe (e.g., system of 2 coupled ODEs), you need to make sure **all** equations are accurately fitted, not just one of them.
- If multiple discovered equations from different tool calls have similar data-fitting errors, choose the candidate with a simpler symbolic form.
    
\end{lstlisting}

\subsection{Tool Implementations and Specifications}
\label{sec:tool-details}

\paragraph{Python code interpreter.}
This tool takes the LLM-generated Python code as its single argument. It evaluates the code and returns all the content printed by the code as the tool output. In the tool specification, we specify that (1) this tool is intended specifically for data analysis tasks only, instead of for symbolic regression itself; (2) the agent should write code to load data files and perform exploratory data analysis, such as computing basic statistics and producing data visualizations when applicable; (3) the code generated by the agent shouold print the analysis results using \texttt{print()} statements, which will be captured and returned as the tool result; (4) when the code intends to save any file to the workspace, such as plots, converted data files, and other non-readable results, it must simultaneously call a predefined \texttt{register\_file()} function to register the file within the workspace, making the file visible to the agent itself at subsequent steps.

\paragraph{Visual subagent.}
We instantiate a GPT-4o-mini LLM subagent specialized in analyzing data visualizations. The tool accepts a single required argument, \texttt{image\_path}, and the following optional arguments:
\begin{itemize}
    \item \texttt{context} (string): Additional context about the data or experiment. Example: \texttt{"This is a physics experiment measuring spring oscillation."} Providing context information helps the vision model provide more relevant insights.
    \item \texttt{focus\_areas} (list of strings): Specific aspects to focus the analysis on, e.g., periodicity, noise level, asymptotic behavior, etc.
\end{itemize}

The system prompt for the subagent is listed as follows.

\begin{lstlisting}[style=sysprompt,caption={System prompt used for the image analysis subagent.},label={lst:system-prompt-image-subagent}]

You are an expert data analyst specializing in visual analysis of scientific plots and data visualizations. Your task is to analyze this plot/visualization and extract insights that would be useful for symbolic regression - the process of discovering mathematical equations that describe the underlying relationship in data.

Analyze the image and provide structured insights in the following categories:

## 1. Observed Patterns
Identify key patterns visible in the data, e.g.:
- Trend direction (increasing, decreasing, constant regions)
- Periodicity or oscillations (estimate frequency/wavelength if visible)
- Growth/decay patterns (linear, exponential, polynomial, logarithmic)

## 2. Potential Functional Forms
Based on visual evidence, suggest likely mathematical forms, e.g.:
- Simple forms: linear (ax+b), polynomial (x^n), power law (x^a)
- Trigonometric (sin, cos)
- Composite forms: products, ratios, nested functions
- For each suggestion, explain the visual evidence supporting it

## 3. Noise Characteristics
Assess data quality:
- Overall noise level
- Noise distribution (uniform, increasing with magnitude, etc.)
- Outliers or anomalies
- Signal-to-noise ratio assessment

## 4. Recommendations for Symbolic Regression
Based on your analysis, provide specific recommendations:
- Which operators to prioritize (Binary: +, -, *, /; Unary: sin, cos, exp, log, sqrt, etc.)
- Template structures to try (e.g., "f(x)*g(y)" for separable effects)

Respond with a JSON object containing your analysis. Example format (you may add or remove fields as needed):
```json
{
  "observed_patterns": {
    "trend": "description of trend or null",
    "periodicity": "description or null",
    "growth_decay": "description or null",
    "asymptotic_behavior": "description or null"
  },
  "potential_functional_forms": [
    {
      "form": "mathematical expression",
      "evidence": "visual evidence description" 
    }
  ],
  "noise_characteristics": "description",
  "sr_recommendations": {
    "binary_operators": ["list", "of", "operators"],
    "unary_operators": ["list", "of", "unary", "ops"],
  }
}
```

\end{lstlisting}

\paragraph{Symmetry discovery.}
We incorporate the methods in \citet{yang2023generative,yang2024symmetry,ko2024learning} as a single tool for identifying Lie symmetry in differential equation systems. As we have mentioned in \cref{sec:tools}, this tool trains a tiny neural network as a surrogate model to predict the derivatives (i.e., the target variables in symbolic regression for differential equations) from the observed variables in the dataset; then, it fixes the surrogate predictor and optimizes for a $q\times q$ matrix $A$ that generates the symmetry group of the trained predictor, where $q$ is the number of variables in the differential equation system. $A$ is optimized under the equivariance error
\begin{equation}
    \mathbb E_{\mathbf x}\|J_{\mathbf f_\theta}(\mathbf x)\cdot ((A\mathbf x)\cdot \nabla)(\mathbf x) - ((A\mathbf x)\cdot \nabla)(\mathbf f_\theta(\mathbf x))\|^2,
    \label{eq:equiv-error}
\end{equation}
where $\mathbf x$ denotes the column vector of $q$ dependent variables, $\mathbf f_\theta$ is the trained predictor, and $J_\cdot$ denotes the Jacobian. The loss is minimized to zero when the surrogate predictor is equivariant with respect to the infinitesimal action of the linear vector field $\mathbf v_A = (A\mathbf x)\cdot \nabla$.

We note a few connections and differences between our implementation and other relevant work in symmetry discovery. \citet{yang2023generative} first formulated the problem of learning invariance/equivariance of functions and approached it by training a GAN where the generator outputs transformations on function graphs and the discriminator is trained to distinguish between the original and the transformed data points. The problem formulation of our tool is similar to \citet{yang2023generative}, but we find the GAN-based discovery method is difficult to converge with the relatively small datasets used in this paper.

Therefore, we choose an approach akin to \citet{ko2024learning}, where the role of the discriminator is replaced by a predictor model, and the symmetry generator (in our case, the matrix $A$) is optimized with a task-specific loss measuring the violation of the symmetry indicated by the generator with respect to the predictor. However, different from the setup in \citet{ko2024learning}, which assumes the knowledge of the ground truth differential equation and uses the exact equation as a perfect predictor, we have to train a surrogate predictor and insert it into the loss. For the systems considered in this paper, we find that a tiny 2-layer MLP is sufficient for producing accurate predictions.

The equivariance error \eqref{eq:equiv-error} is referred from Proposition 3.3 (more specifically, Eq. (5)) of \citet{yang2024symmetry}. It is shown in their proposition that, if the expression inside the norm is zero for all state variables $\mathbf x$, then the one-parameter subgroup generated by $\mathbf v_A$ is exactly the symmetry of the predictor. They have used this as a symmetry regularization term with a fixed $A$ to optimize their equation model, whereas we use the same loss term to optimize for an unknown symmetry generated by $A$.

Finally, we note that the equivariance error in \eqref{eq:equiv-error} is only applicable to \textit{linear} symmetries acting solely on \textit{dependent} variables. This is exactly the setup we specify by the vector field in the form $\mathbf v_A = (A\mathbf x)\cdot\nabla$. More generally, some symmetries of differential equation systems can act with nonlinear functions on the total space containing both independent variables (e.g., the time variable in ODE systems) and dependent variables. We may still use an equivariance error term for discovering those symmetries, but the exact formulation would be different from \eqref{eq:equiv-error} and likely more complicated, depending on the restrictions we want to place on the allowed symmetries. For a more in-depth discussion of the application of general Lie symmetries in symbolic regression, we refer the readers to \citet{yang2025discovering}.

\paragraph{PySINDy.} This tool builds on the PySINDy software \citep{Kaptanoglu2022} and the derivative work of EquivSINDy \citep{yang2024symmetry} with the following arguments exposed to the LLM agent:
\begin{itemize}
    \item \texttt{data\_file} (required): Path to the data file containing the trajectory data of ODE/PDE systems.
    \item \texttt{polynomial\_degree}: The maximal degree of polynomials included in the SINDy function library. Default is 3.
    \item \texttt{derivative\_order}: The maximal order of partial derivatives included in the SINDy feature list. Default is 2.
    \item \texttt{normalize\_columns}: Boolean value indicating whether to normalize feature columns byy $L_2$ norm before regression.
    \item \texttt{threshold}: The threshold for sequential thresholding, controlling the sparsity regularization strength.
    \item \texttt{use\_symmetry}: Boolean value indicating whether to use symmetry to constrain the SINDy parameter space following the method in EquivSINDy \citep{yang2024symmetry}. If true, the \texttt{lie\_generator} argument must also be specified.
    \item \texttt{lie\_generator}: a list or array-like object specifying the matrix generator of the Lie symmetry. Must have the shape $q \times q$ where $q$ matches the number of dependent variables in the dataset.
\end{itemize}

\paragraph{PySR.} The official PySR software \citep{cranmer2023interpretable} offers abundant configurable options for genetic-programming-based SR. In practice, we find that including too many configurable parameters disorients the agent, and the agent tends to try out random combinations of parameters without rigorous reasoning. Therefore, in the tool specification, we instruct the agent to prioritize considering the \texttt{expression\_spec} argument, which allows the agent to propose a template structure of the equation based on its prior knowledge.

The \texttt{expression\_spec} argument shouold be formatted as a dictionary with the following format:
\begin{lstlisting}[style=sysprompt]
{"expressions": [...], "variable_names": [...], "combine": "..."}
\end{lstlisting}

The \texttt{expressions} field should contain placeholder function names. These functions represent unknown sub-expressions that will be fitted by PySR. The \texttt{variable\_names} should contain the feature names from the dataset. The \texttt{combine} field should specify a template string indicating how placeholder functions and known parts (variables, known functional expressions) combine, e.g., \texttt{"F / a + f(F, a)"}, \texttt{"sin(f(x1)) * g(x2, x3)"}, \texttt{"f(x1, x2) + g(x3)"}, etc.

Other tool arguments are listed as follows:
\begin{itemize}
    \item \texttt{input\_file}: path to the data file to perform symbolic regression on.
    \item \texttt{binary\_operators}: Binary operators allowed in building the equation. Default is \texttt{["+", "-", "*", "/"]}.
    \item \texttt{unary\_operators}: Unary operators allowed in building the equation. Common options include $\sin,\cos,\exp,\log,\mathrm{sqrt}$. Default is an empty list.
    \item \texttt{niterations}: Max number of PySR iterations allowed.
    \item \texttt{constraints}: An optional dictionary specifying complexity constraints for specific operators. For example, \texttt{\{"+":(5, 5), "*":(3, 3)\}} restricts the left and right operands of addition to have expression complexity (subtree size) at most 5, and the operands of multiplication to have complexity at most 3.
    \item \texttt{nested\_constraints}: An optional dictionary of constraints on operator nesting. For example, \texttt{\{"sin": \{"cos": 0\}\}} prevents expressions like $\sin(\cos(x))$.
\end{itemize}

Other arguments naturally supported by PySR, such as population size and the number of parallel populations, are not exposed to the agent and set to default to avoid causing distractions for the agent.

%

\section{Experiment Details}

\subsection{Baselines and Common Setups}
\label{sec:baseline-details}

\paragraph{LLM-SR.}

We run LLM-SR~\citep{shojaee2024llm} with the following configurations. We use GPT-4o-mini as the language model backbone to iteratively generate candidate equation skeletons. We allow a maximum of 500 sampled equation proposals for the ODE/PDE datasets during the search process, with each equation constrained to at most 10 free parameters. This is different from the default setting of 2500 samples in their original paper. For the LSR-Transform benchmark, we allow a maximum of 200 samples each dataset and record the wall-clock runtime and token usage.

For the prompts of LLM-SR, we provide an overall description of the domain of each dataset and a list of descriptions of the physical meanings of each variable. We reuse the LLM-SRBench \citep{shojaee2025llm} construction of prompts for the LSR-Transform datasets. For the ODE/PDE systems we collect in \texttt{DiffEq}, we specify the name of each variable and derivative and whether it belongs to the input (feature) variables or the output (target) variables for symbolic regression.

\paragraph{PySR.}

We follow the setup described in \citet{xia2025sr} for evaluating PySR on LLM-SRBench. Specifically, we set the number of iterations to \(125\) and cycles per iteration to \(550\). We also configure populations to \(15\) with a population size of \(33\), a maximum size of \(30\), and a randomization weight of \(0.1\). The set of binary operators used is \(\{ +, -, \times, \div, \mathrm{pow} \}\), and the set of unary operators is \(\{ \exp, \log, \sqrt{\phantom{x}}, \sin, \cos \}\).


We use the complete datasets from LSR-Transform and \texttt{DiffEq} equations for both LLM-SR and PySR, except for the reaction-diffusion system. The training dataset for this system contains the solution to a PDE on a $128\times 128\times 160$ spatiotemporal grid, causing significant computational overhead for evaluating expressions on the full dataset. Therefore, we randomly sample 2000 points from the full dataset before running both baselines.

We evaluate all discovered equations from different methods (including baselines and our method) using a GPT-4o-mini language model with the following prompt:

\begin{lstlisting}[style=sysprompt, caption=Prompt for the LLM-based expression evaluator.]
Given the ground truth mathematical expression A and the hypothesis B, determine if the hypothesis is equivalent to the given ground truth expression up to some error tolerance for the coefficients.

Ground Truth Expression (A): {ground_truth}

Hypothesis Expression (B): {hypothesis}

Let's think step by step. Explain your reasoning and then provide the final answer as:
```json
{{
  "reasoning": "brief step-by-step analysis",
  "answer": "yes/no"
}}
```
\end{lstlisting}

\subsection{Differential Equation Datasets}
\label{sec:diffeq-details}

\begin{table}[ht]
\caption{List of differential equation systems in our \texttt{DiffEq} benchmark.}
    \centering
    {\setlength{\tabcolsep}{10pt}
    \begin{tabular}{cl}
    \toprule
        System name & \multicolumn{1}{c}{Equations} \\
    \midrule
        Bacterial respiration & $\left\{\begin{aligned}
            \dot x &= 20 - x - \frac{xy}{1+0.5x^2} \\
            \dot y &= 10 - \frac{xy}{1 + 0.5x^2}
        \end{aligned}\right.$ \\ [22pt]
        Bar magnets & $\left\{\begin{aligned}
            \dot \theta &= 0.5\sin(\theta - \phi) - \sin(\theta) \\
            \dot \phi &= 0.5\sin(\phi - \theta) - \sin(\phi)
        \end{aligned}\right.$ \\ [20pt]
        Glider & $\left\{\begin{aligned}
            \dot v &= -0.05v^2 - \sin(\theta) \\
            \dot \theta &= v - \frac{\cos(\theta)}{v}
        \end{aligned}\right.$ \\ [22pt]
        Lotka-Volterra & $\left\{\begin{aligned}
            \dot x &= -x^2-2xy+3x \\
            \dot y &= -y^2-xy+2y
        \end{aligned}\right.$ \\ [18pt]
        Predator-prey & $\left\{\begin{aligned}
            \dot x &= x\left(4-x-\frac{y}{1+x}\right) \\
            \dot y &= y\left(\frac{x}{1+x} - 0.075y\right)
        \end{aligned}\right.$ \\ [26pt]
        Shear flow & $\left\{\begin{aligned}
            \dot \theta &= \cot(\phi)\cos(\theta) \\
            \dot \phi &= \sin(\theta)\left(\cos^2(\phi) + 0.1\sin^2(\phi)\right)
        \end{aligned}\right.$ \\ [20pt]
        van der Pol oscillator & $\left\{\begin{aligned}
            \dot x &= 10 \left(y - \frac{1}{3}(x^3 - x)\right) \\
            \dot y &= -\frac{1}{10}x
        \end{aligned}\right.$ \\ [26pt]
        Damped oscillator & $\left\{\begin{aligned}
            \dot x &= -0.1x - y \\
            \dot y &= x - 0.1y
        \end{aligned}\right.$ \\ [20pt]
        Growth & $\left\{\begin{aligned}
            \dot x &= -0.3x + 0.1y^2 \\
            \dot y &= y
        \end{aligned}\right.$ \\ [20pt]
        Reaction-diffusion & $\left\{\begin{aligned}
            u_t &= u(1 - u^2 - v^2) + v(u^2+v^2) + 0.1(u_{xx} + u_{yy}) \\
            v_t &= v(1 - u^2 - v^2) - u(u^2+v^2) + 0.1(v_{xx} + v_{yy})
        \end{aligned}\right.$ \\ [10pt]
    \bottomrule
    \end{tabular}
    }
    \label{tab:diffeq-desc}
\end{table}

\cref{tab:diffeq-desc} lists the 10 differential equation systems we consider for evaluation in this paper. Among them, the first 7 systems come from \citet{strogatz2001nonlinear}, a textbook for nonlinear dynamical systems. \citet{la2021contemporary} proposes to generate data from these systems as a benchmark for SR methods and has incorporated these datasets into the Penn Machine Learning Benchmark (PMLB) \citep{olson2017pmlb}. In our experiments, we use their existing datasets for these systems, each containing 4 solution trajectories of the same ODE system with 100 timesteps.

In addition, we include three systems used in the experiments of \citet{yang2024symmetry}: damped oscillator, growth, and reaction-diffusion. We follow their data generation procedure to construct our dataset. The damped oscillator dataset contains 50 trajectories, each with 20 timesteps and an interval $\Delta t = 0.2$. The growth dataset contains 10 trajectories, each with 100 timesteps and an interval $\Delta t = 0.02$. For these two datasets, we split them on the trajectory dimension into 80\%:20\% for training and testing, respectively. The reaction-diffusion system is originally used in \citet{champion19} for testing equation discovery methods. It is governed by two PDEs, posing additional challenges as the partial derivatives inflate the feature space for symbolic regression. We reuse the data provided by \citet{champion19}, which contains one solution trajectory as a $128 \times 128 \times 200$ array ($128\times 128$ spatial grid and $200$ timesteps). We split the data along the time dimension, using the first $160$ timesteps for training and the remaining $40$ steps for testing.

To construct the noisy \texttt{DiffEq} datasets, we add white noise to each state variable, whose scale $\sigma_i$ depends on the variance of each state variable in the dataset: $\sigma_i = \sigma_R\cdot \mathrm{std}(x_i)$. We refer to the coefficient $\sigma_R$ as the noise level. Then, the targets (time derivatives) are estimated by applying the central difference method to the noisy trajectories. For all systems except reaction-diffusion, we use a noise level of $0.01$. For reaction-diffusion, we choose a noise level of $0.0001$, as a larger noise level, together with the inclusion of partial derivatives in this problem, makes it challenging for all methods considered in this paper to discover the correct equations.

\section{Additional Results}

\subsection{Long-Term Prediction in \texttt{DiffEq} Systems}
\label{sec:ltp-diffeq}

\begin{figure}[ht]
    \centering
    \includegraphics[width=\linewidth]{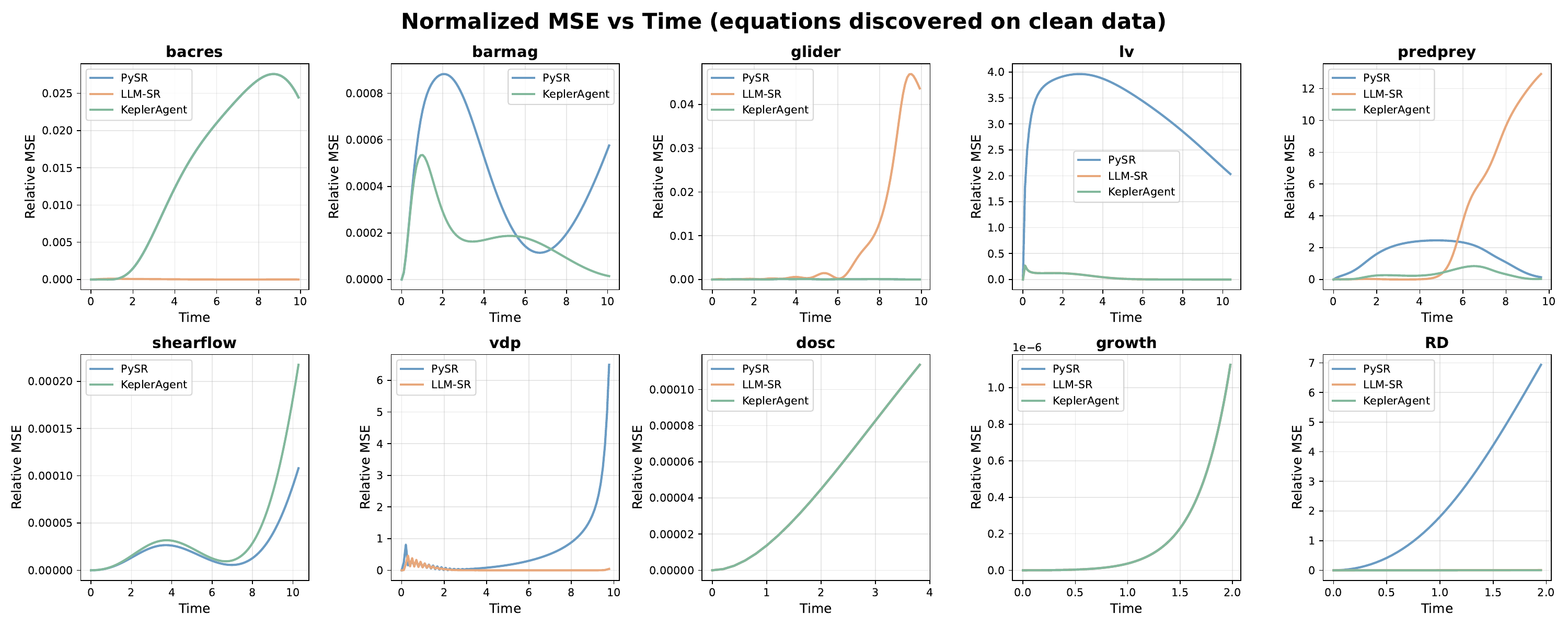}
    \caption{Plotting normalized MSE against time for equations discovered on clean data from PySR, LLM-SR, and our KeplerAgent. A method is not included in a subplot if its discovered equation causes simulation failure (e.g., state variables going to infinity, or other numerical issues).}
    \label{fig:ltp-diffeq-clean}
\end{figure}

\begin{figure}[ht]
    \centering
    \includegraphics[width=\linewidth]{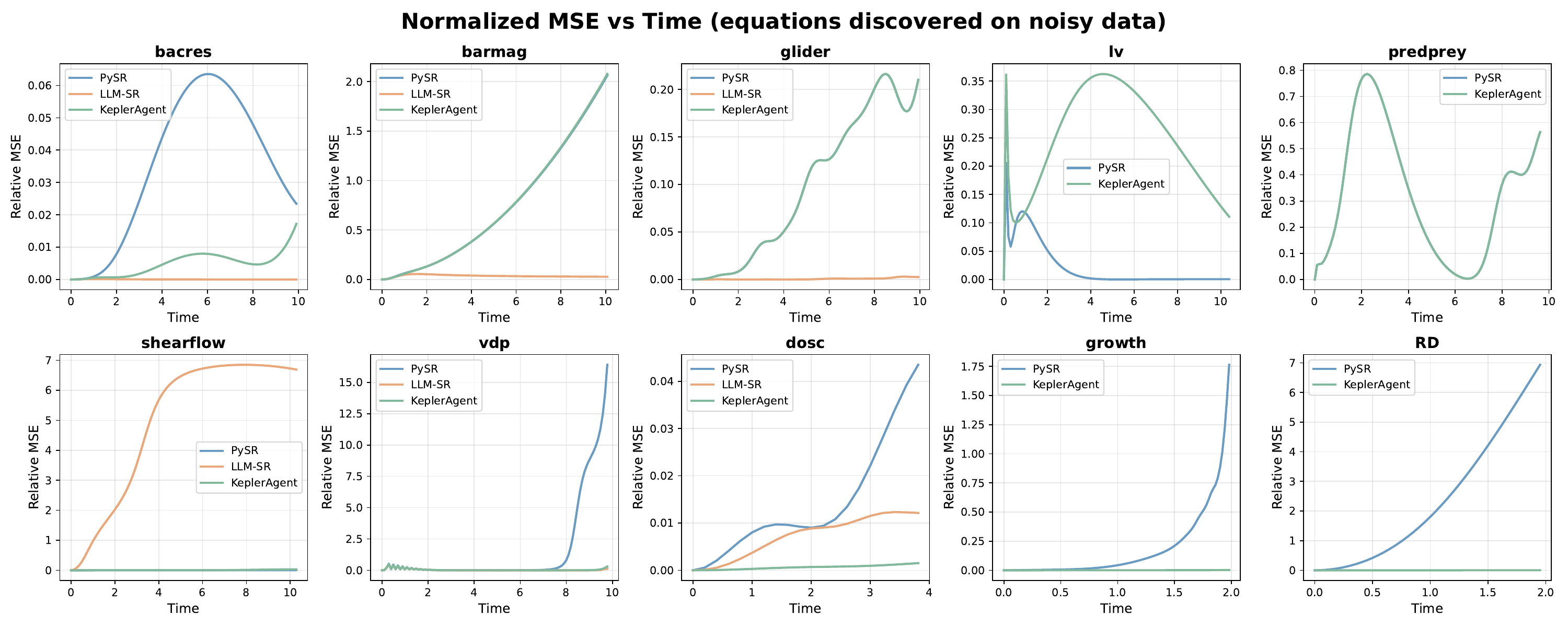}
    \caption{Plotting normalized MSE against time for equations discovered on noisy data from PySR, LLM-SR, and our KeplerAgent.}
    \label{fig:ltp-diffeq-noisy}
\end{figure}

In addition to the results in \cref{tab:diffeq}, \cref{fig:ltp-diffeq-clean} and \ref{fig:ltp-diffeq-noisy} show the full error growing curves when we use the discovered equations from different methods for long-term prediction. Specifically, in each subplot, we start from an initial point in a trajectory from the test dataset and use a numerical ODE solver to integrate the discovered equations until the maximum time available in the dataset. Then, the simulated trajectories are compared with the true trajectory in the dataset at each available time step. For equation discovery on the clean data, we observe that KeplerAgent maintains relatively low errors throughout the simulation time span in almost all systems (except in the van der Pol oscillator, where the simulation failed). In comparison, PySR and LLM-SR suffer from high prediction errors in systems like glider, Lotka-Volterra, and predator-prey, and LLM-SR causes simulation failure twice on the shear flow system and the bar magnets system.

In the noisy data regime, KeplerAgent remains relatively robust, performing best or close to best among the three methods in 6 of the 10 systems (predator-prey, shear flow, van der Pol oscillator, damped oscillator, growth, and reaction-diffusion).

\subsection{An Instance of KeplerAgent Reasoning Trace}

In this subsection, we inspect a real execution of KeplerAgent on the reaction-diffusion system (described in \cref{tab:diffeq-desc}) to understand how KeplerAgent reasons and orchestrates different tools to obtain the final equation discovery result. Instead of showing the full outputs of the LLM agent, we provide snapshots of the experience log at each step, interleaved with key parts of the reasoning performed by the agent, which clearly represent the agent's journey in this discovery problem.

In the first step, the agent learns that the dataset contains a solution trajectory to a PDE. Therefore, it needs to run a symbolic regression tool that is suitable for discovering differential equations from data, and it identifies PySINDy as a good choice. The agent has also considered using symmetry constraints if relevant symmetries can be discovered from the dataset. However, since it doesn't have any prior knowledge about the symmetries, it starts with a straightforward SINDy approach with default parameters.

\begin{lstlisting}[style=sysprompt]
## Experience Log

### Step 1: sindy
**Arguments:**
  - data_file: rd_train.h5
**Result:**
  - Status: success
  - result_type: equation
  - equations: [
    '0.041*u + 0.035*u*u*u + 1.169*v*v*v + 0.007*u*v*v + 0.277*u*u*v + 0.071*u_yy + 0.065*u_xx',
    '0.041*v - 1.169*u*u*u + 0.035*v*v*v - 0.277*u*v*v + 0.007*u*u*v + 0.065*v_yy + 0.071*v_xx'
    ]
  - mape: 70.033%
\end{lstlisting}

Given that the equations from the first PySINDy call have not achieved satisfactory fitting (MAPE $= 70\%$), the agent decides to proceed with the following steps: (1) Since the system is governed by PDEs, discovering symmetries could provide valuable insights into the governing equations. This might help in constraining the subsequent symbolic regression process. (2) After discovering symmetries, rerun the SINDy algorithm with the symmetry constraints to see if it improves the fitting of the equations. The next tool call for symmetry discovery reflects the first step in this plan.

\begin{lstlisting}[style=sysprompt]
### Step 2: symmetry_discovery
**Arguments:**
  - data_file: rd_train.h5
**Result:**
  - Status: success
  - result_type: symmetry
  - lie_generator: [[0.0, -0.678013801574707], [0.680248498916626, 0.0]]
  - predictor_loss: 0.009965
  - symmetry_loss: 0.002001
\end{lstlisting}

The agent notices that the symmetry discovery step provided a Lie generator matrix indicating some underlying symmetries in the system. Both the predictor loss and the symmetry loss were low, which suggests that the system has some structured behavior that can be exploited. Given that the SINDy results did not yield satisfactory equations, and a symmetry discovery result is available, which can help refine the equation search, the agent decides to proceed with the SINDy tool again, incorporating the symmetry constraints obtained from the previous step.

Notably, the agent notices that the Lie generator result in the last tool call is close to the exact form of a rotation generator $[[0, -1]; [1, 0]]$. Given that the direct tool result is obtained from numerical optimization and may have some error, and that it is close to a commonly observed structure (rotation symmetry) in the real world, the agent decides to proceed with the exact rotation generator. We comment that this reasoning step is crucial and fills the gap between existing methods for symmetry discovery and symmetry-constrained equation modeling. Existing symmetry discovery methods often rely on gradient-based optimization, which produces numerical results that slightly deviate from the exact solution. Previously, research in symmetry-constrained equation discovery, such as \citet{yang2024symmetry}, often relies on humans in the loop to inspect discovered symmetry manually and translate it into an exact solution that makes sense physically, e.g., the rotation generator above. However, in this reasoning step, we show that KeplerAgent is capable of interpreting the numerical results from these tools and making the most reasonable guess about the true answer based on its prior knowledge, thereby fulfilling the entire pipeline from discovering symmetry, processing and interpreting the result, to using this intermediate finding in the downstream equation discovery task, without requiring any human intervention. While we have only incorporated a few tools in the current agent and observed this ability explicitly only for symmetry discovery, we believe that the agent also has the potential to apply this kind of reasoning ability flexibly with other tools.

\begin{lstlisting}[style=sysprompt]
### Step 3: sindy
**Arguments:**
  - data_file: rd_train.h5
  - use_symmetry: True
  - lie_generator: [[0, -1], [1, 0]]
**Result:**
  - Status: success
  - result_type: equation
  - equations: [
    '1.008*u - 1.017*u*u*u + 0.999*v*v*v - 1.017*u*v*v + 0.999*u*u*v + 0.094*u_yy + 0.094*u_xx',
    '1.008*v - 0.999*u*u*u - 1.017*v*v*v - 0.999*u*v*v - 1.017*u*u*v + 0.094*v_yy + 0.094*v_xx'
    ]
  - mape: 15.584%
\end{lstlisting}

In this step, with the symmetry constraint enabled, the agent manages to obtain a list of equations with much lower MAPE. However, it is still much above the specified threshold of $0.1\%$ (because of the noise in the data, which the agent is not aware since it has chosen not to call the code execution tool for data analysis in the beginning). Therefore, for the next step, the agent makes another attempt to call the PySINDy tool, but with a slightly different configuration.

\begin{lstlisting}[style=sysprompt]
### Step 4: sindy
**Arguments:**
  - data_file: rd_train.h5
  - use_symmetry: True
  - lie_generator: [[0, -1], [1, 0]]
  - polynomial_degree: 4
**Result:**
  - Status: success
  - result_type: equation
  - equations: [
    '1.008*u - 1.017*u*u*u + 0.999*v*v*v - 1.017*u*v*v + 0.999*u*u*v + 0.094*u_yy + 0.094*u_xx',
    '1.008*v - 0.999*u*u*u - 1.017*v*v*v - 0.999*u*v*v - 1.017*u*u*v + 0.094*v_yy + 0.094*v_xx'
    ]
  - mape: 15.584%
\end{lstlisting}

After step 4, the agent discovers that changing the polynomial degree parameter in SINDy yields the same result. As the error is still above the threshold by a large amount, the agent starts some repetitive attempts to call SINDy with different configurations, e.g., further increasing the polynomial degree, adjusting threshold values, etc, in the hope that the result can be further improved. However, the current MAPE turns out to be the best the agent can obtain from existing tools. In the end, the agent keeps calling tools until the maximum number of iterations (10) is reached, and then returns the discovered equations in step 4 as the final result.

We comment that, while the first few reasoning steps are impressive, the later steps with repetitive tool calls (which we omit here) reveal further room for improvement. For example, the agent should learn more from the previous experience: the fact that increasing the polynomial degree leads to the same equations likely means that a low polynomial degree already suffices for this task, and that further increasing the degree is not helpful at all. Also, the agent needs to be better at navigating the entire toolkit, e.g., making use of the code interpreter to perform noise analysis of the data, or using PySR to explore the possibility of more complex equation forms, instead of sticking to the same tool until the maximum computation budget. We identify these issues here as potential directions for future work.


\end{document}